\documentclass[journal]{IEEEtran}
\usepackage{cite}
\usepackage{times}
\usepackage{epsfig}
\usepackage{graphicx}
\usepackage{amsmath}
\usepackage{amssymb}
\usepackage{multirow}
\usepackage{longtable}
\usepackage{color}
\usepackage{rotating}
\usepackage{makecell}
\usepackage{array}
\usepackage{booktabs}
\usepackage{colortbl}
\usepackage{arydshln}
\usepackage{mathrsfs}
\usepackage{amsmath}
\usepackage{float}
\usepackage{caption}
\usepackage{subcaption}
\usepackage{amssymb}
\usepackage{bbm}
\graphicspath{{figures/}}

\begin{document}
\title{Collaborative Representation for SPD Matrices with Application to Image-Set Classification}

\author{Li Chu,
        Rui Wang,
        and~Xiao-Jun Wu*~\IEEEmembership{}% <-this % stops a space
\thanks{L. Chu, R. Wang, and X.-J. Wu (\textit{Corresponding author}) are with the School of Artificial Intelligence and Computer Science, Jiangnan University, Wuxi 214122, China. L. Chu, R. Wang, and X.-J. Wu are also with Jiangsu Provincial Engineering Laboratory of Pattern Recognition and Computational Intelligence, Jiangnan University e-mail: (15861899728@163.com; cs\_wr@jiangnan.edu.cn; xiaojun\_wu\_jnu@163.com).}
}

\maketitle

\begin{abstract}
Collaborative representation-based classification (CRC) has demonstrated remarkable progress in the past few years because of its closed-form analytical solutions. However, the existing CRC methods are incapable of processing the nonlinear variational information directly. Recent advances illustrate that how to effectively model these nonlinear variational information and learn invariant representations is an open challenge in the community of computer vision and pattern recognition To this end, we try to design a new algorithm to handle this problem. Firstly, the second-order statistic, i.e., covariance matrix is applied to model the original image sets. Due to the space formed by a set of nonsingular covariance matrices is a well-known Symmetric Positive Definite (SPD) manifold, generalising the Euclidean collaborative representation to the SPD manifold is not an easy task. Then, we devise two strategies to cope with this issue. One attempts to embed the SPD manifold-valued data representations into an associated tangent space via the matrix logarithm map. Another is to embed them into a Reproducing Kernel Hilbert Space (RKHS) by utilizing the Riemannian kernel function. After these two treatments, CRC is applicable to the SPD manifold-valued features. The evaluations on four banchmarking datasets justify its effectiveness. 
\end{abstract}


\IEEEpeerreviewmaketitle

\section{Introduction}
\IEEEPARstart {S}parse representation based classification (SRC) has recently attracted increasing attention in the domain of digital image processing and pattern recognition due to its robustness \cite{zhang2011sparse,yang2011fisher,nestfuse,li2017multi,luo2017image,li2011no,sang2014no,kristan2020eighth,sun2019effective}. It finds a sparse representation of the testing sample over the dictionary with sparsity constraints computed by minimizing the $L_1$-norm based reconstruction error. Based on this, many papers explore innovative applications or design more discriminative dictionary learning models to further improve the visual classification performance \cite{yang2011fisher,qiao2010sparsity,zhang2012kernel}. Recently, some works have shown that collaborative representation is more efficient than $L_{1}$-norm based sparsity constraint. To this end, Zhang et al.\cite{Lei2011Sparse} propose a collaborative representation based classification (CRC) method using the $L_{2}$-norm based regularization, and the experimental results show that CRC is able to achieve competitive classification performance but with lower computational complexity than SRC. Hence, CRC has attracted much attention due to its effectiveness \cite{zhang2012collaborative,li2014joint,cai2016probabilistic}.

However, the existing collaborative representation methods are mainly designed for Euclidean visual data, which can not encode the nonlinear variational information between image sets efficiently. Recently, image set classification has been proven to be an area of increasing vitality in the computer vision and pattern recognition community \cite{huang2015log,chen2018Component,wu2015learning,chen2018riemannian,cheng2017duplex,wang2018simple,wang2017discriminative,Wang2018Multiple,wang2020multiple,wang2021geometry}. Compared to single-shot image-based classification \cite{wu2004new,zheng2006nearest,chen2018new,shen2013content,luo2016novel,feng2017face}, each image set generally contains a large number of images that belong to the same class and cover large variations such as pose changes, illumination differences and partial occlusions. When referring to the mathematical models of image set data, covariance matrix \cite{wang2012covariance,harandi2014manifold,harandi2018dimensionality}, linear subspace \cite{hamm2008grassmann,huang2015projection,wang2020graph} and Gaussian distribution\cite{wang2015discriminant} are three well-known representation methods, among which, covariance matrices provide a natural data representation by computing the second-order statistic of a group of samples. Thus, we select it to describe each image set in this paper. According to the previous works \cite{huang2015log,wang2017discriminative,wang2012covariance}, we know that the space formed by a family of covariance matrices with the same dimentionality is a non-linear Riemannian manifold. Specifically, it is an SPD manifold. As a consequence, the Euclidean learning approaches can not be applied to the SPD manifold-valued data directly. To handle this problem, some distance metrics have been developed on the Riemannian manifold for similarity measurement, such as Log-Euclidean Metric (LEM) \cite{arsigny2007geometric} and Affine-Invariant Riemannian Metric (AIRM) \cite{pennec2006riemannian}. By making use of these Riemannian metrices, the Euclidean computations can be generalized to the non-Euclidean data representations according to the following tactics.

{\color{black}Inspired by the superiority of kernel representation \cite{zheng2006reformative}, the first type of tactic is to embed the original Riemannian manifold into a Reproducing Kernel Hilbert Space (RKHS) via Riemannian kernel functions \cite{wang2012covariance,hamm2008grassmann,harandi2012sparse,jayasumana2013kernel}. Although these methods can achieved favourable performance on several visual classification tasks, the Riemannian geometrical structure of the original set data may be distorted in the process of RKHS transformation. To solve this issue, some algorithms have been put forward to jointly perform similarity metric learning and embedding mapping learning on the original Riemannian manifolds directly. Accordingly, a lower-dimensional and more discriminative subspace can be generated for image set classification. Compared with the aforementioned strategy, this type of approach takes the Riemannian geometry of the original data manifold into fully consideration in the dimensionality reduction process, thus may lead to impressive classification performance. As a matter of fact, the methods mentioned above just model each given image set from a single geometrical perspective, which may lose some useful structural information for visual data. As a countermeasure, Wang et al.\cite{Wang2018Multiple} and Huang et al. \cite{huang2015face} utilize the metric learning method to learn an effective metric space for the heterogeneous features extracted from the multiple Riemannian manifold-valued descriptors.} 

\begin{figure}[!t]
\centering
  \includegraphics[width=\linewidth]{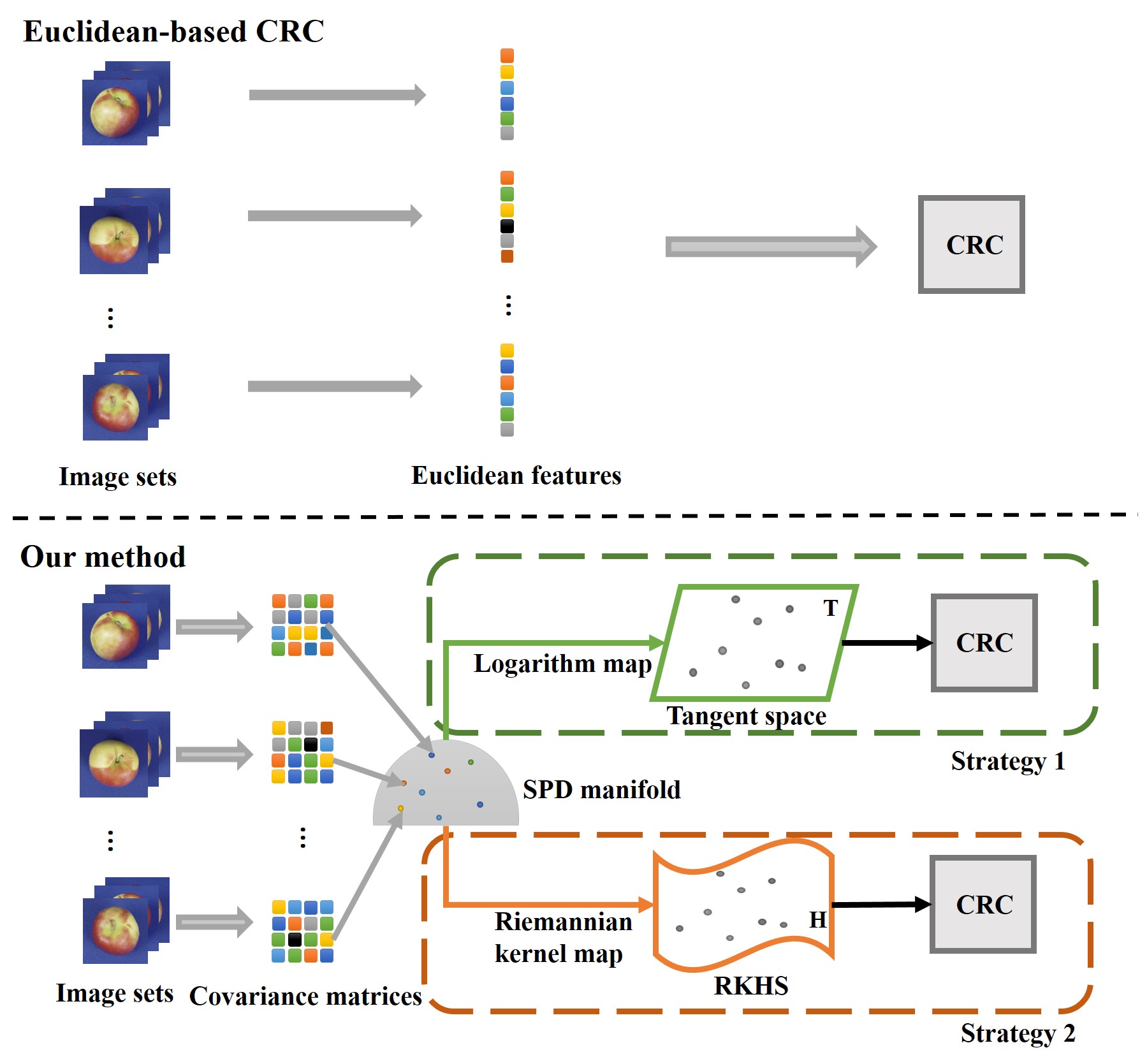} 
  \caption{Overview of the Euclidean-based CRC and our method. Euclidean-based CRC: Each image set is described as a vector feature. Our method: Each image set is represented by a covariance matrix which resides on the SPD manifold. To make collaborative representation work for the manifold-value data, we provide two strategies, one is to embed the SPD manifold into a tangent space via matrix logarithm map(Strategy 1 in the figure). Another transforms the SPD manifold to RKHS by exploiting Riemannian kernel function(Strategy 2 in the figure).}
  \label{fig1}
\end{figure}

In parallel with the above developments, sparse representation methods have received widespread attention in the context of Riemannian manifold due to its simplicity and striking performance \cite{Kai2010Action,sra2011generalized,harandi2012sparse,li2013log,harandi2016sparse}. Obviously, these methods make use of sparse representation to perform representation learning, and $L_{1}$-norm based sparsity usually leads to higher computational cost. Since the efficient $L_{2}$-norm based collaborative representation has already shown its discriminatory power and higher computational efficiency, it is valuable to study whether CRC can also perform well on the SPD manifold. This study is planned for proposing the first attempt in this domain. With this objective, we present a collaborative representation-based image set classification in the scenario of SPD manifold in this paper. Fig. \ref{fig1} indicates the basic idea of traditional Euclidean-based CRC and our approach. Compared with the Euclidean-based CRC (the top line of Fig. \ref{fig1}), we use the covariance matrix to model each image set. Generally speaking, the space formed by a set of covariance matrices is not a vector space but instead adhering to an SPD manifold. In order to make the collaborative representation applicable to the SPD manifold, we offer two strategies. The first strategy is to embed the original SPD manifold into an associated tangent space via matrix logarithm map (we use Log\_CRC to represent this method). Another strategy is to exploit the Riemannian kernel function to map the original SPD manifold into RKHS (we call this method LogEK\_CRC). In these two cases, the collaborative representation can be applied to conduct further feature learning and classification. The evaluations on four different benchmarking datasets verify the effectiveness of the proposed method.
%%%%%%%%%%%%%%%%%picture%%%%%%%%
%%%%%%%%%%%%%%%%%%%%%%%%%%%%%%%%%%

\section{Related Works}

{\color{black}The existing Riemannian manifold-based image set classification methods can be grouped into four categories, such as Riemannian kernel-based discriminative learning methods \cite{wang2012covariance, hamm2008grassmann,harandi2012sparse,jayasumana2013kernel,wang2020graph}, Riemannian manifold dimensionality reduction methods \cite{huang2015log,harandi2014manifold,harandi2018dimensionality,huang2015projection}, multiple statistics fusion-based methods \cite{Wang2018Multiple,huang2015face}, and Riemannian manifold-based sparse representation methods \cite{Kai2010Action,sra2011generalized,harandi2012sparse,li2013log,harandi2016sparse}. The working mechanism of the first type of approach is to embed the original Riemannian manifold into a Reproducing Kernel Hilbert Space (RKHS) via Riemannian kernel functions  \cite{wang2012covariance,hamm2008grassmann,harandi2012sparse,jayasumana2013kernel}. Therein, Wang et al. \cite{wang2012covariance} derive a Log-Euclidean Metric-based kernel function to transform the original SPD manifold into RKHS.  Similarly, Hamm et al. \cite{hamm2008grassmann} embed the original Grassmann manifold into RKHS via the projection kernel function. Besides, both \cite{wang2012covariance} and \cite{hamm2008grassmann} utilize the kernel discriminant analysis algorithm to perform discriminant subspace learning. In order to improve the description ability of the Riemannian kernel-based approach, Wang et al. \cite{gemkml} first builds a lightweight feature extraction network on the original Grassmann manfiold for the sake of learning fine-grained geometric features, Then, the projection kernel function is applied to map the generated data representations into the kernel space for the subsequent metric learning process. However, one shortcoming of this type of approach is that the Riemannian geometrical structure of the original set data may be changed in the feature transformation process. 

To solve this problem, a slice of algorithms have been proposed to jointly perform similarity metric learning and embedding mapping learning on the original Riemannian manifolds directly. Wherein, Harandi et al. \cite{harandi2018dimensionality} propose to model the orthonormal mapping from the original high-dimensional SPD manifold to a lower-dimensional one with a metric learning framework. Similarly, Huang et al. \cite{huang2015projection} construct a metric learning framework on the original Grassmann manifold for the purpose of directly yielding a lower-dimensional and more appropriate Grassmann manifold for classification. Differently, Huang et al. \cite{huang2015log} propose a novel Log-Euclidean metric learning framework on the SPD manifold tangent space, which can not only produce a desirable new SPD manifold, but also demonstrate higher computational efficiency. As a matter of fact, the aforementioned approaches just model each given image set from a single geometrical perspective, which may lose some useful structural information for visual classification. Therefore, Wang et al. \cite{Wang2018Multiple} and Huang et al. \cite{huang2015face} propose to model each given image set with different visual statistics simultaneously. Since different statistics reside in different topological spaces, the metric learning framework is desinged to fuse these hybrid features into a unified subspace for improved classification. 

In parallel with the above developments, the Riemannian manifold-based sparse representation methods have obtained widespread attention due to its simplicity and efficiency in representation learning \cite{Kai2010Action,sra2011generalized,harandi2012sparse,li2013log,harandi2016sparse}. Therein, Kai et al.\cite{Kai2010Action} propose to embed the original SPD manifold-valued data into an associated tangent space via matrix logarithm map. In this case, the existing sparse representation methods apply. Different from \cite{Kai2010Action}, Harandi et al.\cite{harandi2012sparse} utilize the Stein kernel to embed the original SPD manifold-valued features into RKHS. Consequently, the sparse representation also can be carried out in it. Similarly, Li et al.\cite{li2013log} and Harandi et al.\cite{harandi2016sparse} try to generalise the sparse representation to the Riemannian manifolds by exploiting the Log-Euclidean kernel and Jeffrey kernel, respectively.} 

\section{Background Theory}
\indent Before introducing our method, we briefly review the covariance matrix, tangent space, and Riemannian kernel function, respectively.
\subsection{Covariance matrix}
\indent Let $S_{i}=\left \{ s_{1},s_{2},...,s_{n_{i}} \right \}$ be an image set with $n_{i}$ images, where $s_{i}\in\textup{R}^{d}$ is the $i-$th vectorized instance. We model the image set $S_{i}$ as the $d \times d$ covariance matrix\cite{wang2012covariance},\cite{ harandi2014manifold},\cite{harandi2018dimensionality}:
\begin{equation}
\label{cov}
X =\frac{1}{n-1}\sum_{i=1}^{n}\left ( s_{i}-u_{S_i} \right )\left ( s_{i}-u_{S_i} \right )^{T}
\end{equation}
where $u_{S_{i}}$ is the mean of $S_i$. In order to prevent covariance matrix from being singular, we add a small perturbation value to $X$ : $ \hat{X}\Leftarrow\mathop X+\lambda I$, where $\lambda$ is set to $10^{-3}\times trace\left ( C \right )$ and $I$ is the identity matrix\cite{wang2012covariance}.

\subsection{Tangent space and Riemannian kernel function}
\indent The covariance matrix $X$ computed by Eq.(\ref{cov}) does not lie in a vector space but instead adhering to a nonlinear Riemannian manifold, i.e., SPD manifold $Sym_\textup{d}^+$. As a consequence, the Euclidean computations are inapplicable. Arsigny et al.\cite{arsigny2006log} propose to use the matrix logarithm map to transform the manifold of covariance matrices into a vector space of symmetric matrices. Let $X=U \Sigma U^{T}$ be the eigen-decomposition of covariance matrix $X$, where $\Sigma$ is the diagonal matrix of eigenvalues, $U$ is the corresponding eigenvectors matrix. Then, the matrix logarithm map can be expressed as 
\begin{equation}
\label{log2}
\textup{log}\left ( X \right )=U\textup{log}\left ( \Sigma  \right )U^{T}
\end{equation}
where $\textup{log}\left ( \Sigma  \right )$ is the diagonal matrix of the eigenvalue logarithms.

\indent When it comes to similarity measurement on SPD manifold, Log-Euclidean Metric(LEM)\cite{arsigny2007geometric} is a widely used distance measure, which corresponds to an Euclidean metric in the logarithmic domain. The distance between two points $X_i, X_j$ on $Sym_\textup{d}^+$ is computed by LEM as:
\begin{equation}
\label{LEM}
d_{LEM}\left ( X_i,X_j \right )=\left \| \textup{log}\left ( X_i \right )-\textup{log}\left ( X_j \right ) \right \|_\textup{F}
\end{equation}
where $\left \|\cdot   \right \|_\textup{F}$ means the matrix Frobenius norm.\\
\indent Under this metric, a Riemannian kernel\cite{li2013log} can be formulated as:
\begin{equation}
\label{LE_K}
k\left ( X_i,X_j \right )=\textup{exp}\left ( -\beta \left \| \textup{log}\left ( X_i \right )-\textup{log}\left ( X_j \right ) \right \|_\textup{F}^{2} \right )
\end{equation}
the validity of which has been confirmed in\cite{li2013log}.

\section{Collaborative representation for SPD Matrices}
\indent This section minutely presents the proposed collaborative representation method for SPD matrices. Specifically, Section $\textup{A}$ discusses the strategic details of mapping the SPD manifold into an associated tangent space, and we introduce the collaborative representation in RKHS in Section $\textup{B}$. Let $\widetilde S =\left \{ S_{1},S_{2},...,S_{N}\right \}$ be the gallery composed by $N$ image sets, with each set expressed as: $S_{i}=\left \{ s_{1},s_{2},...,s_{n_{i}} \right \}\in \textup{R}^{d\times n_{i}}$, where $i=1\to N$, and $n_{i}$ means the number of images in the $i-$th image set, $S_{i}$ belongs to category $c_i$. From the previous studies\cite{wang2012covariance},\cite{ harandi2014manifold},\cite{harandi2018dimensionality}, we know that each image set can be represented by a covariance matrix in Eq.(\ref{cov}). Then, we use $\widetilde X=\left \{ X_1,X_2,...,X_N\right \}$ to denote the computed SPD manifold-valued feature representations of the gallery $\widetilde S$.
% embedding by Riemannian kernel function

\subsection{Collaborative representation in tangent space(Log\_CRC)}
\indent Collaborative representation has been proven to have a good performance for Euclidean data. But, it can not be applied to the SPD manifold-valued features directly. To address this problem, we embed each sample of $\widetilde X$ into the tangent space by using the logarithm map $\Psi _\textup{log}:Sym_\textup{d}^+\rightarrow T$, where $T$ means the associated tangent space of $\widetilde X$. And the points in the tangent space are expressed as $L_X=\left \{ L_1, L_2,......,L_N \right \}, L_{i}=\textup{log}(X_{i})\in \textup{R}^{d\times d}$. To facilitate the subsequent computations, we further convert each tangent map into a vector. $l_X=\left \{ l_1, l_2,......,l_N\right \}$ is used to denote the vectorized matrix of $L_X$ and $l_{i}=vec (L_{i})\in \textup{R}^{d^{2}}$. Similarly, we can use the same procedure to obtain the vectorized query set. For each query data $y$, it can be collaboratively represented as $l_X$ and classified by measuring which class leads to the minimum reconstruction error. The representation vector $w$ is obtained by solving the following objective function: 
\begin{equation}
\label{vec-crc}
w=\textup{min}\left \{ \left \| y-l_X w  \right \|_{2}^{2}+\lambda_{1} \left \| w \right \|_{2} \right \}
\end{equation}
where $\lambda_{1}$ is the regularization parameter. 

\indent The solution to Eq.\ref{vec-crc} can be easily derived as:
\begin{equation}
\label{vec-crc2}
w=\left ( l_X^{T}l_X+\lambda_{1} I \right )^{-1}l_X^{T}y
\end{equation}
where $I$ is an identity matrix. The reconstruction error between the query sample $y$ and the gallery with respect to class $c$ is
\begin{equation}
\label{err-crc}
Err_{c}=\left \|y -l_X^{c}w_{c} \right \|_{2}/\left \| w_{c} \right \|_{2}
\end{equation} 
where $l_X^{c}$ denotes the data collected by $l_X$ from the $c-$th class and $w_{c}$ is the corresponding collaborative representation coefficients with respect to class $c$, $y$ is assigned to the class with the smallest reconstruction error
\begin{equation}
\label{label}
label=\underset{c}{\textup{min}}\left ( Err_{c} \right ),c=1,2,...k
\end{equation}
where $k$ indicates the number of classes.

%%%%%%%%%%%%%%%%%%%%%%%%%
%Algorithm
\begin{table} [t]
\renewcommand\arraystretch{1.3}
\centering
\scriptsize 
\resizebox{3.5in}{!}{
\begin{tabular}{m{6cm}} 
\hline 
$\textbf{Algorithm:}$ The proposed method\\
\hline
$\textbf{Input:}$ \\
$\bullet$ The gallery: $\widetilde X=\left \{ X_1,X_2,...,X_N\right \}$, $X_i\in Sym_\textup{d}^+$\\
$\bullet$ The regularization parameter: $\lambda_{1}$, $\lambda_{2}$\\
$\textbf{Output:}$ \\
$\bullet$ Class label \\
\hline
$\textbf{Log\_CRC method:}$\\
1: Get the tangent space data $L_X=\left \{ L_1, L_2,......,L_N \right \}$ by logarithm map in Eq.$(\ref{log2})$.\\
2: Obtain the vectorized data in the tangent space: $l_X=\left \{ l_1, l_2,......,l_N\right \}$.\\
3: Compute the collaborative representation coefficient for each vectorized test point $y$ via: \\
$w=\left ( l_X^{T}l_X+\lambda_1 I \right )^{-1}l_X^{T}y$.\\
4: Compute the reconstruction error of $y$ and the data for class $c$ via $Err_{c}=\left \| y -l_X^cw_{c} \right \|_{2}/\left \| w_{c} \right \|_{2}$.\\
5: Get the class label by $\underset{c}{\textup{min}}\left ( Err_{c} \right ),c=1,2,...k$.\\
\hline
$\textbf{LogEK\_CRC method:}$\\
1: Compute the $K_{\widetilde X\widetilde X}$, $K_{\widetilde XY}$ and $K_{YY}$ by Riemannian kernel function in Eq.(\ref{LE_K}).\\
2: Perform SVD on $K_{\widetilde X\widetilde X}=U\Sigma U^{T}$.\\
2: Compute $\Psi = \left ( U\Sigma ^{1/2} \right )^{T}$ in Eq.(\ref{svd}) and then get $\Phi = \left ( \Psi ^{T} \right )^{-1}K_{\widetilde XY}$.\\
4: Get the collaborative representation coefficient for each query point $Y\in Sym_\textup{d}^+$ via:\\
 $w=\left ( \Psi^{T}\Psi+\lambda_{2} I \right )^{-1}\Psi^{T}\Phi$.\\
5: Compute $\underset{c}{\textup{min}}\left \{ \left \|\Phi-\Psi_{c}w_{c} \right \|_{2}/\left \| w_{c} \right \|_{2} \right \}$ to get the class label of $Y$.\\
\hline
\end{tabular}}
\end{table}
%%%%%%%%%%%%%%%%%%%%%%%%%%
%%%%%

\subsection{Collaborative representation in RKHS(LogEK\_CRC)}
\indent To make the Euclidean collaborative representation applicable to the SPD manifold-valued data, another strategy is to transform the SPD manifold to RKHS by  employing the Riemannian kernel function specified in Eq. \ref{LE_K}. This mapping process is expressed as: $\o:Sym_\textup{d}^+\rightarrow H$. We use $\o(\widetilde X)= \left \{ \phi (X_1), \phi (X_2),......, \phi (X_N)\right \}$ to represent the new feature representation of $\widetilde X$ in Riemannian kernel space $H$. For each query point $Y$ on $Sym_\textup{d}^+$, the Riemannian kernel collaborative representation\cite{wang2013kernel} in the RKHS is formulated as:
\begin{equation}
\label{KCRC}
\begin{split}
\resizebox{.65\hsize}{!}{$w=\textup{min}\left \{ \left \| \o \left ( Y \right )-\o(\widetilde X)w \right \|_{2}^{2} +\lambda_{2} \left \| w \right \|_{2}^{2}\right \}$}
\end{split}
\end{equation}
where $\lambda_{2}$ is the regularization parameter. According to Mercer theorem\cite{gao2012sparse}, the inner product between two points in kernel space can be represented by their kernel function $k(X_{i},X_{j})=\left \langle \o \left ( X_{i} \right ),\o \left ( X_{j} \right ) \right \rangle$. With the Riemannian kernel defined in Eq.$(\ref{LE_K})$, Eq.$(\ref{KCRC})$ is expanded as
\begin{equation}
\begin{split}
\label{KCRC2}
\resizebox{.7\hsize}{!}{$K_{YY}-2w^{T}K_{\widetilde XY}+w^{T}K_{\widetilde X\widetilde X}w+\lambda_{2} \left \| w \right \|_{2}^{2} $}
\end{split}
\end{equation}
where, $K_{\widetilde X\widetilde X}$ is a $N\times N$ matrix, each element of the matrix describes the similarity between data in the gallery, $i.e$,  $k(X_i,X_j)$, where $i=1,2,...N$ and $j=1,2,...N$. $K_{\widetilde XY}$ is a $N\times 1$ vector which consists of $k(X_i,Y), i=1,2,...N$.\\
\indent Inspired by \cite{wu2015manifold}, in order to further simplify Eq.$(\ref{KCRC2})$, the matrix $K_{\widetilde X\widetilde X}$ is rewritten through Singular Value Decomposition (SVD)
\begin{equation}
\label{svd}
\begin{split}
\resizebox{.7\hsize}{!}{$K_{\widetilde X\widetilde X}=U\Sigma U^{T}=U\Sigma ^{1/2}\left ( \Sigma ^{1/2} \right )^{T}U^{T}=\Psi ^{T}\Psi $}
\end{split}
\end{equation}
where $\Sigma$ is the diagonal matrix of eigenvalues, $U$ is the corresponding eigenvectors matrix. $\Psi = \left ( U\Sigma ^{1/2} \right )^{T}$ is the approximate representation of $\widetilde X$ in Riemannian kernel space, i.e., $\o(\widetilde X)$. Further, $K_{\widetilde XY}$ can be  rewriten as
\begin{equation}
\label{svd2}
\resizebox{.6\hsize}{!}{$K_{\widetilde XY}=\Psi ^{T}\left ( \Psi ^{T} \right )^{-1}K_{\widetilde XY}=\Psi ^{T}\Phi $}
\end{equation}
where $\Phi = \left ( \Psi ^{T} \right )^{-1}K_{\widetilde XY}$ is the approximate representation of $Y$ in Riemannian kernel space, i.e., $\o(Y)$. Similarly, $K_{YY}=\Phi ^{T}\Phi $. Thus, we get
\begin{equation}
\begin{split}
\label{KCRC3}
&\Phi ^{T}\Phi - 2w^{T}\Psi ^{T}\Phi + w^{T}\Psi ^{T}\Psi w+\lambda_{2} \left \| w \right \|_{2}^{2}\\
&=\left \| \Phi -\Psi w \right \|_{2}^{2}+\lambda_{2} \left \| w \right \|_{2}^{2}
\end{split}
\end{equation}
$w$ can be derived as
\begin{equation}
\label{vec-crc2}
w=\left ( \Psi^{T}\Psi+\lambda_{2} I \right )^{-1}\Psi^{T}\Phi
\end{equation}
then we obtain the label of $Y$ 
\begin{equation}
\label{label-kcrc}
\begin{split}
label&=\underset{c}{\textup{min}}\left \{ \left \| \phi (Y)-\phi (\widetilde X)_{c}w_{c} \right \|_{2}/\left \| w_{c} \right \|_{2} \right \}\\
&=\underset{c}{\textup{min}}\left \{ \left \|\Phi-\Psi_{c}w_{c} \right \|_{2}/\left \| w_{c} \right \|_{2} \right \}\\
& c=1,2,...,k 
\end{split}
\end{equation} 
where $\Psi_{c}$ is the data associated with the $c-th$ class, $w_{c}$ is the collaborative representation coefficients with respect to class $c$, $k$ indicates the number of classes.
%%%%%

\subsection{Relation With the Previous Works}
{\color{black}The propsoed approach is related to two previous works \cite{WeiD, ZhouP}. In this part, we summarize some essential differences between our method and those introduced in \cite{WeiD} and \cite{ZhouP} in the following two paragraphs.

\textit{Relation With \cite{WeiD}}. In fact, both the proposed algorithm and \cite{WeiD} try to generalise the Euclidean collaborative representation to the Riemannian manifold-valued feature representations for the sake of improving the image set classification performnce on some challenging visual scenarions. However, there exist some fundamental differences among them: 1) for feature representation, the propsoed approach is executed in the scenario of SPD manifiold, while \cite{WeiD} is carrried out in the context of Grassmann manifold; 2) based on the generated Grassmann manifold-valued features, \cite{WeiD} constructs the $P+V$ model to encode the inter-class similarity information and the intra-class variational information of the data, while the propsoed approach does not consider this explicitly; 3) with the constructed $P+V$ model, \cite{WeiD} implements the collaborative representation framework with prototype learning in the space spanned by a set of symmetric matrices. As a result, the dictionary learning problem could be converted into standard Euclidean problem. Different from \cite{WeiD}, our method implements it in two ways. The first is to map the original SPD manifold into an associated tangent space, and the second is to embed it into the RKHS. Since both the tangent space and the RKHS conform to the Euclidean geometry, the collaborative representation applies.

\textit{Relation With \cite{ZhouP}}. Actually, both the proposed algorithm and \cite{ZhouP} devote to extend the traditional single image-based collaborative representation to the image set-based one to inject new vigour for image set classification. But, they demonstrate different representation learning mechanisms for this problem. Specifically, for similarity measurement between the query set and each gallery set, \cite{ZhouP} model the query set and the whole training dictionary as a convex hull to define the collaborative representation-based set-to-sets distance for classification. Hence, \cite{ZhouP} has two characteristics: 1) the distinctiveness of the images in the query set has been exploited; 2) the feature learning and classification process is implemented in the Euclidean space. Different from \cite{ZhouP}, our approach first model the original training and test image sets onto the SPD manifold for the purpose of mining the nonlinear geometrical structural information of the data. In order to perform similarity measurement and classification, we offer two tactics. One is to map the SPD manifold-valued features into the tangent space, another is to transform them into RKHS. In these two scenarios, the Euclidean collaborative representation framework can be naturally applied. However, the proposed method does not consider the distinctiveness of the instances in the query set in the representation learning process. The aforementioned differences also leads to the different objective functions among them. Besides, the propsoed approach is evaluated on four different visual classification tasks, while \cite{ZhouP} is limited to face recognition task.  
}

\section{Experiments And Analysis}
\indent We test the classification performance of our method on four typical visual tasks: object categorization, face recognition, virus cell classification and dynamic scene classification. For the task of set-based object categorization, we use the ETH-80 dataset\cite{leibe2003analyzing}. The widely used YouTube Celebrities (YTC) dataset \cite{wang2012covariance} is applied to the face recognition task. We utilize the MDSD dataset\cite{shroff2010moving} for the task of video-based dynamic scene classification and Virus \cite{kylberg2011virus} for virus cell classification task.
%%%%%%%%%%%%%%%%%%%%%%%%%%%%%%%%%
\begin{table*}[!ht]
\renewcommand\arraystretch{1.4}
\centering

\caption{ \upshape Average recognition rates and standard deviations of different menthods on ETH-80 \cite{leibe2003analyzing}, Virus \cite{kylberg2011virus} ,  MDSD \cite{shroff2010moving} and YTC\cite{wang2012covariance} datasets.}
\label{ACC}
%\LARGE
\tiny
\resizebox{4.8in}{!}{
\begin{tabular}{ccccc}
\hline 
Method 	   &ETH-80\cite{leibe2003analyzing}&Virus \cite{kylberg2011virus}&MDSD \cite{shroff2010moving}&YTC\cite{wang2012covariance}  \\
\hline
CDL\cite{wang2012covariance}				&93.75$\pm$3.43	      &48.33$\pm$3.60     &30.51$\pm$2.82      &68.72$\pm$2.96\\
GDA\cite{hamm2008grassmann}		     &93.25$\pm$4.80      &47.00$\pm$2.49     &30.51$\pm$7.78      &65.78$\pm$3.34\\
PML\cite{huang2015projection}		          &90.00$\pm$3.53      &47.33$\pm$3.43     &29.32$\pm$4.66      &67.62$\pm$3.32\\
LEML\cite{huang2015log}		                     &92.25$\pm$3.19      &55.67$\pm$9.94     &29.74$\pm$3.89      &69.04$\pm$3.84 \\
MMML\cite{Wang2018Multiple}               &95.00$\pm$1.89      &51.13$\pm$7.67     &31.95$\pm$6.26      &76.70$\pm$2.81 \\
SPDML-AIRM\cite{harandi2014manifold}	 &91.65$\pm$3.34    &52.33$\pm$8.42     &21.21$\pm$5.06      &64.66$\pm$2.92\\
SPDML-Stein\cite{harandi2014manifold}		&90.83$\pm$3.62     &51.60$\pm$8.10      &21.79$\pm$4.88      &61.87$\pm$3.32\\
\hline
Frob\_SRC\cite{sra2011generalized}	      &93.25$\pm$4.43      &54.75$\pm$7.83     &23.75$\pm$5.67      &63.16$\pm$2.72\\
LogEK\_SRC\cite{li2013log}		                 &94.88$\pm$4.17      &56.33$\pm$6.37     &27.95$\pm$4.94      &72.55$\pm$2.77 \\
Log\_SRC\cite{Kai2010Action}                    &95.12$\pm$4.17      &55.67$\pm$7.04     &34.83$\pm$5.70      &74.68$\pm$2.71 \\
$\textbf {LogEK\_CRC}$                               &96.50$\pm$3.76      &61.67$\pm$3.93     &35.64$\pm$4.09      &78.87$\pm$2.44 \\
$\textbf {Log\_CRC}$		                           &$\textbf {97.50}$$\pm$1.67   &$\textbf {64.67}$$\pm$5.02      &$\textbf {36.15}$$\pm$5.19     &$\textbf {78.90}$$\pm$2.53   \\
\hline
\end{tabular}}
\end{table*}
%%%%%%%%%%%%%%%%%%%%%%%%%%%%%%%%%%%%%%%%%%%%%%%%%%%%%%%%%%%%%%

\subsection{Datasets and settings} 
\begin{figure}[!t]
\centering
  \includegraphics[width=0.7\linewidth]{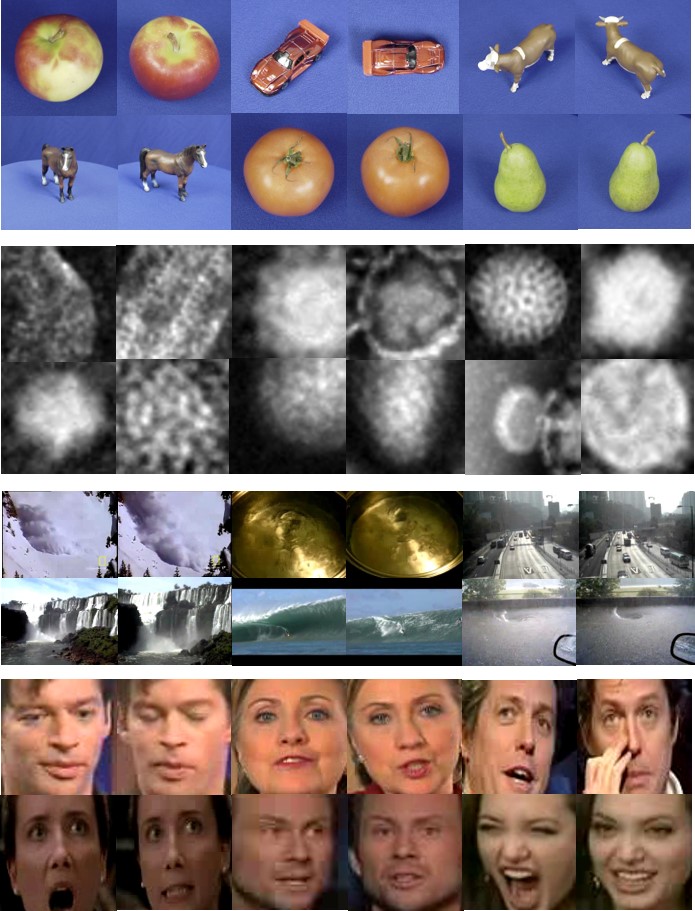} 
  \caption{Examples in four datasets. Top line: ETH-80. The second line: Virus. The third line: MDSD. Bottom line: YTC.  }
    \label{fig-data}
\end{figure}

\indent The ETH-80 dataset consists of 8 categories, such as apples, cows, cups, dogs, horses, pears, tomatoes and cars, with each category has 10 image sets, and each image set contains 41 images of different perspectives.We randomly select five from each category for traning and the remaining five for testing. The top line of Fig. \ref{fig-data} shows some examples of this dataset.

\indent The Virus dataset is composed of 15 categories, each of which consists of 5 image sets. There are 20 different images in each set. For this dataset, 3 image sets are randomly chosen for training and the rest 2 image sets are used for testing. Some examples of Virus dataset are presented in the second line of Fig. \ref{fig-data}.

\indent The MDSD dataset contains 13 different categories of dynamic scenes, with each class consists of 10 video sequences collected in an unconstrained setting. We randomly select 7 videos for training and the rest for testing in each class. As presented in the third line of Fig. \ref{fig-data}, there are some sample images of this dataset.

\indent For the YouTube Celebrities dataset, there are 1910 video clips of 47 subjects involved in it. Each clip consists of hundreds of face frames. We randomly choose 9 video clips in each subject with 3 for training and 6 for testing.  Some sample face frames of this dataset are shown in the bottom line of Fig. \ref{fig-data}.
%%%%%%%%%%%%%%%%%picture%%%%%%%%
%%%%%%%%%%%%%%%%%%%%%%%%%%%%%%%%%%

\indent To keep consistent with the previous works\cite{wang2012covariance} \cite{Wang2018Multiple} \cite{huang2015log}	\cite{Kai2010Action} \cite{li2013log}, we conduct ten-fold cross-validation experiments, i.e., repeat the randomly selecting process of the gallery/probes ten times. Then, the average classification scores are reported for each method on the four used datasets. Besides, we resize each given image into a $20\times20$ grayscale one. 

%%%%%%%%%%%%%%%%%picture%%%%%%%%
\begin{figure*}[ht]
\centering
  \includegraphics[width=1.0\linewidth]{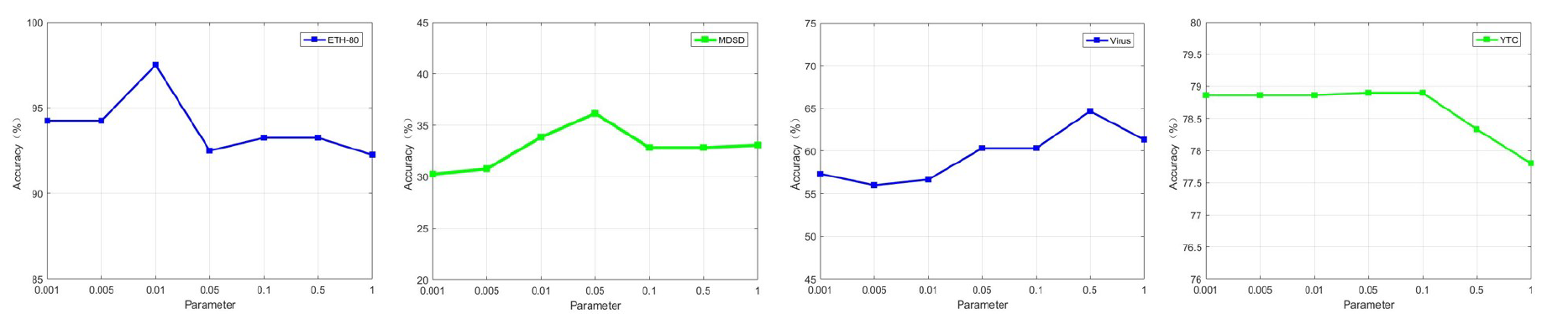} 
  \caption{The effect of regularization parameter $\lambda_{1}$ on accuracy for Log\_CRC method.  }
  \label{fig-par1}
\end{figure*}
%%%%%%%%%%%%%%%%%%%%%%%%%%%%%%%%%%
%%%%%%%%%%%%%%%%%picture%%%%%%%%
\begin{figure*}[ht]
\centering
  \includegraphics[width=1.0\linewidth]{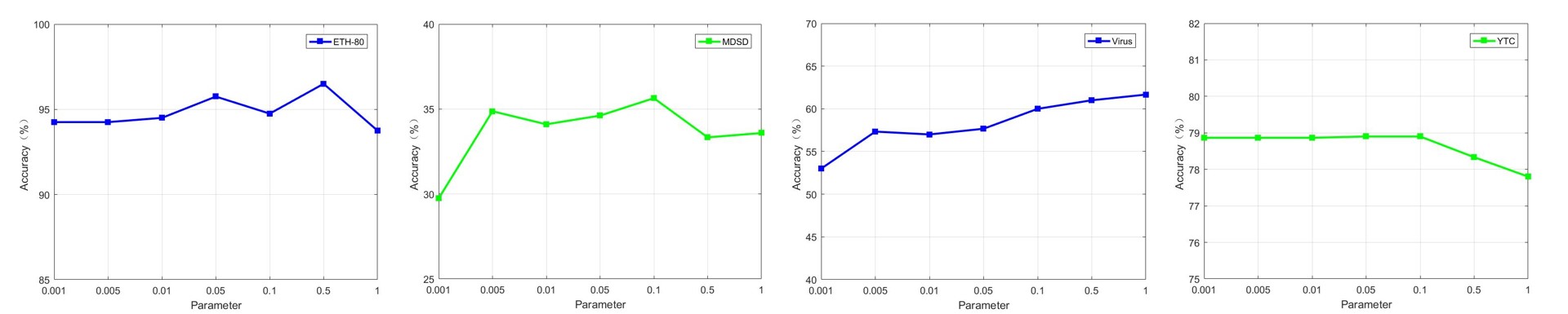} 
  \caption{The effect of regularization parameter $\lambda_{2}$ on accuracy for LogEK\_CRC method.  }
  \label{fig-par2}
\end{figure*}
%%%%%%%%%%%%%%%%%%%%%%%%%%%%%%%%%%
\subsection{Comparative methods}
\indent In order to study the effectiveness of the proposed method, we compare the proposed method with some representative image set classification methods including: Covariance Discriminant Learning(CDL)\cite{wang2012covariance} , Grassmann Discriminant Analysis (GDA)\cite{hamm2008grassmann}, Projection Metric Learning (PML)\cite{huang2015projection}, Log-Euclidean Metric Learning (LEML)\cite{huang2015log}, Multiple Manifolds Metric Learning(MMML)\cite{Wang2018Multiple}, SPD Manifold Leaning (SPDML-AIRM, SPDML-Stein)\cite{harandi2014manifold}, Generalized Dictionary Learning and Sparse Coding using Frobenius Norm(Frob\_SRC)\cite{sra2011generalized}, Logarithm Mapping for Sparse Representation(Log\_SRC)\cite{Kai2010Action}, and Log-Euclidean Kernels for Sparse Representation(LogEK\_SRC)\cite{li2013log}

\indent We should emphasize that we use the source codes of all the comparative methods provided by the original authors to conduct experiments on the four datasets. For CDL, the perturbation is set to $10^{-3}\times trace\left ( X \right )$. For GDA, the number of basis vectors for the subspace are determined by crossvalidation. In PML, the number of iterations, the trade-off coefficient $\alpha$ and the target dimensionality $d$ are reported according to the original work. In LEML, the parameter $\eta$ is tuned in the range of $\left [ 0.1,1,10 \right ]$, $\zeta$ is tuned from $0.1$ to $0.5$. For SPDML, following the original work\cite{harandi2014manifold}, $v_{w}$ is fixed as the minimum number of samples in one class while the dimensionality of the target SPD manifold is tuned by cross-validation. For LogEK\_SRC, we use Riemannian kernel function in Eq.(\ref{LE_K}).

\subsection{Result and Analysis}
\indent Table \ref{ACC} lists the recognition rates of different comparative methods on the four datasets. Among them, we can intuitively see that our method, especially Log\_CRC, exhibits better performance in terms of recognition scores on the four datasets. It is also interesting to see that the recognition rates of PML are lower than that of LEML. The main difference of them is that PML conducts the projection metric learning on the Grassmann manifold, while LEML learns the Log-Euclidean Metric on the SPD manifold, which justifies the superiority of LEM-based SPD manifold dimensionality reduction. The classification performance comparison between SPDML-Stein and SPDML-AIRM further verifies the effectiveness of the LEM-based Riemannian metric learning frameworks for image set classification. Obviously, MMML performs better than other competitors on the ETH-80, MDSD, and YTC datasets, because of the complementarity of multiple Riemannian manifolds is qualified to mine more powerful structural information for improved classification.

\indent Secondly, compared with the sparse representation methods Frob\_SRC,  Log\_SRC, LogEK\_SRC and collaborative representation methods Log\_CRC, LogEK\_CRC, we can find that collaborative representation-based classification in tangent space and Riemannian kernel space for SPD matrix perform better than sparse representation-based classification. This could demonstrate that the collaborative representation is more efficient than $L_{1}$-norm based sparse constraint. 

\indent Due to the regularization parameters $\lambda_{1}$ and $\lambda_{2}$ play a pivotal role in our method, we utilize cross-validation to choose suitable values for them on the four used datasets. Fig. \ref{fig-par1} and Fig. \ref{fig-par2} illustrate their impact on the classification results of Log\_CRC and LogEK\_CRC. According to these two figures, the best value of $\lambda_{1}$ on the ETH-80, Virus, MDSD and YTC datasets are set to 0.01, 0.5, 0.05, and 0.1, respectively. For $\lambda_{2}$, 0.5, 1, 0.1, and 0.05 are the selected proper values for the ETH-80, Virus, MDSD, and YTC datasets, respectively.

\begin{table}[!t]
\renewcommand\arraystretch{1.2}
\centering
\caption{The average classification scores of the different methods on the ETH-80, Virus, MDSD, and YTC datasets.}
\label{tab-ec}
\tiny
\resizebox{3.4in}{!}{
\begin{tabular}{ccccc}
\hline 
Method 	   &ETH-80 &Virus &MDSD &YTC\\
\hline
$\textbf {LogEK\_CRC}$     &96.50$\pm$3.76      &61.67$\pm$3.93     &35.64$\pm$4.09      &78.87$\pm$2.44 \\
$\textbf {Log\_CRC}$		  &$\textbf {97.50}$$\pm$1.67   &$\textbf {64.67}$$\pm$5.02      &$\textbf {36.15}$$\pm$5.19     &$\textbf {78.90}$$\pm$2.53   \\
CRC   &91.05$\pm$3.94      &24.67$\pm$2.81     &32.05$\pm$6.86      &63.69$\pm$3.61 \\
SPD\_CRC & 95.50$\pm$2.84 & 40.00$\pm$6.85 & 34.62$\pm$5.13  & 64.96$\pm$2.89 \\
\hline
\end{tabular}}
\end{table}

%%%%%%%%%%%%%%%%%%%%%%%%%%%%%%%%%
\begin{table}[ht]
\renewcommand\arraystretch{1.2}
\centering
\caption{ \upshape Testing time(s) comparison of five methods on ETH-80 dataset.(For classifying one image set.)}
\label{time}
%\LARGE
\tiny
\resizebox{2.0in}{!}{
\begin{tabular}{cc}
\hline 
Method 	   &ETH-80\cite{leibe2003analyzing}  \\
\hline
Frob\_SRC\cite{sra2011generalized}	      &0.212\\
LogEK\_SRC\cite{li2013log}		                 &0.147 \\
Log\_SRC\cite{Kai2010Action}                    &0.259 \\
$\textbf {LogEK\_CRC}$                                  &0.127 \\
$\textbf {Log\_CRC}$		                           &$\textbf {0.078} $\\
\hline
\end{tabular}}
\end{table}
%%%%%%%%%%%%%%%%%%%%%%%%%%%%%%%%%%%%%%%%%%%%%%%%%%%%%%%%%%%%%%

\subsection{Ablation Study for Each Component of the Proposed Framework}
{\color{black}Although the effectiveness of the proposed approach has been demonstrated, to validate the contribution of each component of the proposed framework is also highly appealing. To this end, we make a comparison between LogEK\_CRC, Log\_CRC, CRC, and SPD\_CRC on the ETH-80, Virus, MDSD, and YTC datasets to observe their difference in classification performance. Here, 'CRC' represents that we directly exploit the Euclidean collaborative representation to perform feature selection and classification for the original image set samples. For 'SPD\_CRC', it first models the original set data onto the SPD manifold, then makes use of the Euclidean collaborative representation to conduct feature learning and classification in this space. The final classification results of them are listed in Table \ref{tab-ec}. According to this table, it is evident that the classification accuracies of CRC are significantly lower than that of other three competitors on the four used datasets. The reason may be that the original image set samples contain large within-class variational information, which restrains CRC from mining more pivotal information for classification. After modeling these data onto the SPD manifold, the classification ability of SPD\_CRC has been further promoted compared to CRC, which confirms the effectiveness of Riemannian manifold in mining the nonlinear geometrical structure of the visual data. However, an obvious limitation of SPD\_CRC is the Euclidean collaborative representation-based classification is directly carried out on the SPD manifold, which inevitably leads to the sub-optimal solution. From Table \ref{tab-ec}, it is also worth noting that the proposed LogEK\_CRC and Log\_CRC outperform SPD\_CRC and CRC in terms of classification score on the four used datasets, visibly. These experimental observations demonstrate the significance of each module of the proposed framework for image set classification.} 

\subsection{Testing Time Comparison}
\indent Table \ref{time} presents the testing time of the proposed methods and several sparse representation methods on the ETH-80 dataset. The experiments were run on 3.0GHz PC with 4GB RAM and Matlab2016a software. We need to emphasize that the testing time is computed by classifying one test sample with the whole gallery. From Table \ref{time}, it is clear to see that the testing burden of the proposed Log\_CRC and LogEK\_CRC has been reduced compared to other competitors. The reasons are twofold: 1) our method classifies the query set with the gallery directly instead of classifying by learning a dictionary in SRC methods; 2) our method has closed-form analytical solutions.

\section{Conclusion}
\indent This paper proposes a collaborative representation-based image set classification algorithm in the context of SPD manifold. For this approach, we suggest two ways to implement feature extraction and classification, one is to map the SPD manifold-valued data representations into the tangent space via matrix logarithm map, another is to embed them into RKHS using the Riemannian kernel function. After that, the merits of Euclidean collaborative representation are qualified to be generalised to the SPD manifold.

\indent Experimental results obtained on four different benchmarking datasets show that the proposed approach could make a considerable improvement in classification performance and computational efficiency compared to some representative image set classification methods. For future work, we plan to integrate dictionary learning into the proposed framework for further improving the discriminability of the learned geometric features. 

\bibliographystyle{unsrt} 
\bibliography{Ref}

\end{document}